\documentclass[sigconf]{acmart}

\usepackage{booktabs}
\usepackage{makecell}
\usepackage{framed}

\AtBeginDocument{%
  }

\setcopyright{acmlicensed}
\copyrightyear{2026}
\acmYear{2026}
\acmDOI{10.1145/3820755.3821486}
\acmConference[DocEng '26]{the 26th ACM Symposium on Document Engineering}{August 25--28, 2026}{Fribourg, Switzerland}
\acmISBN{978-1-4503-XXXX-X/2026/XX}

\makeatletter
\renewcommand\@ACM@checkaffil{%
  \if@ACM@instpresent\else
    \ClassWarningNoLine{\@classname}{No institution present for an affiliation}%
  \fi
  \if@ACM@citypresent\else
    \ClassWarningNoLine{\@classname}{No city present for an affiliation}%
  \fi
  \if@ACM@countrypresent\else
    \ClassWarningNoLine{\@classname}{No country present for an affiliation}%
  \fi
}
\makeatother

\begin{document}

\title{Revising RVL-CDIP: Quantifying Errors and Test-Train Overlap}

\author{Stefan Larson}
\affiliation{\institution{Vanderbilt University}}

\author{Attila Nagy}
\affiliation{\institution{ML Collective}}

\author{Sam Desai}
\affiliation{\institution{University of Michigan}}

\author{Cyrus Desai}
\affiliation{\institution{University of Michigan}}

\author{Nicole C. Lima}
\affiliation{\institution{University of Michigan}}

\author{Yixin Yuan}
\affiliation{\institution{University of Michigan}}

\author{Siddharth Betala}
\affiliation{\institution{IIT Madras, ML Collective}}

\author{Kaushal K. Prajapati}
\affiliation{\institution{ML Collective}}

\author{Jamiu T. Suleiman}
\affiliation{\institution{ML Collective}}

\author{Sharad Duwal}
\affiliation{\institution{ML Collective}}

\author{Kevin Leach}
\affiliation{\institution{Vanderbilt University}}

\renewcommand{\shortauthors}{Larson et al.}
\settopmatter{authorsperrow=4}

\begin{abstract}
RVL-CDIP is a popular dataset for benchmarking document classifiers.
However, the dataset contains ample amounts of label errors as well as non-trivial amounts of test-train overlap, both of which may impact model performance metrics.
In this paper, we address these two problems by (1) finding and fixing label errors, and (2) detecting and addressing test-train overlap.
We produce several variations of RVL-CDIP with label error and test-train overlap fixes, and benchmark document classification performance on these new RVL-CDIP variations.
Our rigorous analysis of RVL-CDIP finds that the corpus contains 12\% label error and approximately 35\% test-train duplication.
Remediation sees improvements in classification accuracy when errors are removed, but sees decreases in accuracy when duplicates are removed.
We additionally evaluate models on RVL-CDIP-N, an out-of-distribution benchmark, finding that training on error-corrected data substantially improves OOD generalization, with supervised models gaining an average of 8.1 percentage points in accuracy and improvements as large as 14 percentage points.
\end{abstract}




\maketitle

\section{Introduction}

\begin{figure}[h]
    \centering\scalebox{0.52}{
    \includegraphics{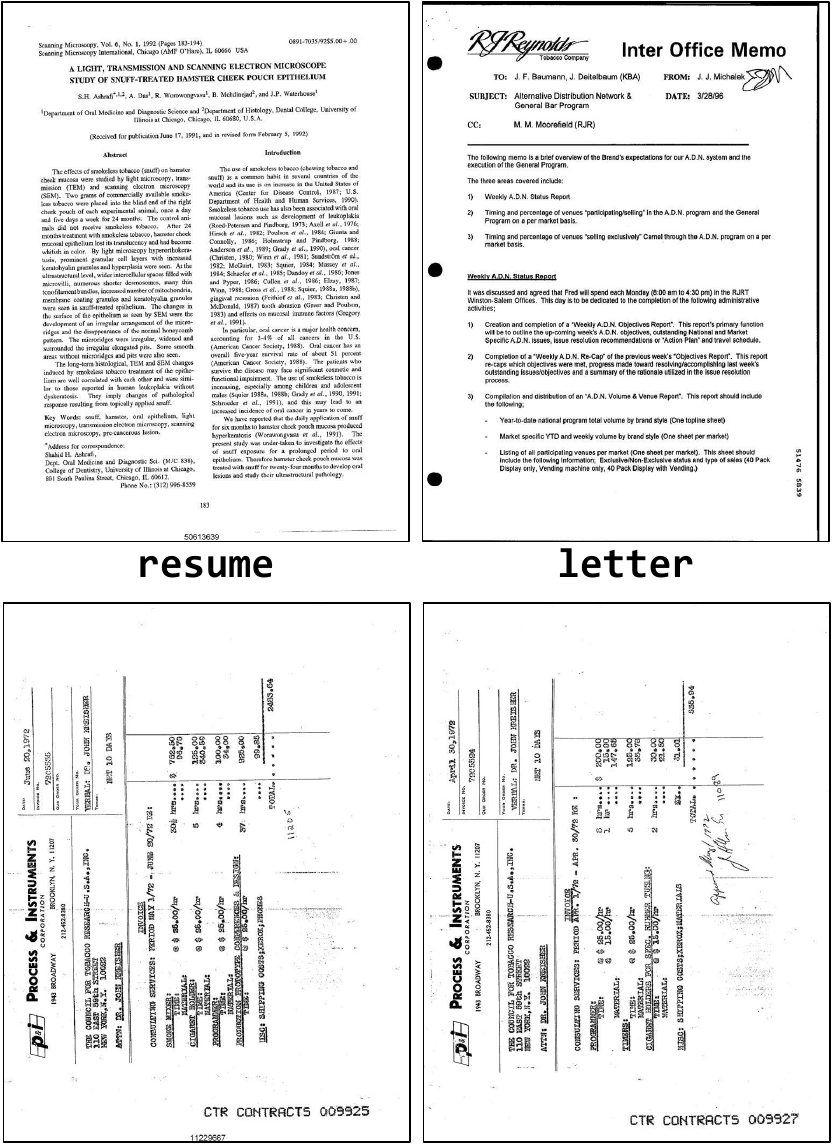}}
    \caption{Example label errors (top row) and test-train near-duplicate pair (bottom row, from the \texttt{invoice} category). The \texttt{resume} document's correct label is \texttt{scientific\_publication}, and the \texttt{letter}'s is \texttt{memo}.}
    \label{fig:teaser}
\end{figure}

The RVL-CDIP document classification corpus~\cite{harley2015icdar-rvlcdip} has been called the \emph{de facto} standard benchmark for evaluating document classification models~\cite{larson-etal-2023-evaluation}.
Indeed, modern document understanding models like LayoutLMv3~\cite{layoutlmv3-huang-2022}, Donut~\cite{kim2022donut}, and UDOP~\cite{tang-2023-udop}, among others, are often exclusively benchmarked on RVL-CDIP to establish performance scores for document classification.
Despite this, recent work has estimated that there are large amounts of label errors in the RVL-CDIP dataset, as well as large amounts of ``data leakage'' --- duplicate and near-duplicate samples --- across RVL-CDIP's test and train splits~\cite{larson-etal-2023-evaluation} (see Figure~\ref{fig:teaser} for examples of these phenomena).
These two undesirable features cast doubt on high reported performance scores, and motivate a rigorous investigation into the impacts of widespread label errors and test-train overlap on models trained and evaluated on RVL-CDIP.

\begin{figure*}
    \centering
    \scalebox{0.43}{
    \includegraphics{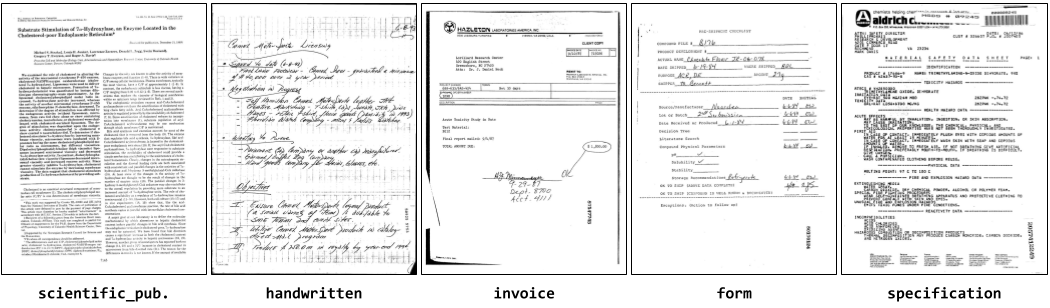}}
    \caption{Example documents from a selection of RVL-CDIP's 16 categories.}
    \label{fig:rvlcdip-examples}
\end{figure*}

In this paper, we seek to identify and fix the label errors found in RVL-CDIP, and we aim to detect and address the data leakage observed across the dataset's test and train splits.
We manually review RVL-CDIP and uncover a large amount of label errors: our analysis finds that roughly 12\% of RVL-CDIP is erroneously labeled, with error rates in the \texttt{letter} category exceeding 30\%.
We then apply a filter-and-refine method to detect test-train overlap, finding that roughly 35\% of RVL-CDIP's test set contains a (near-) duplicate counterpart in the dataset's train set.
We remediate these data quality issues and create several ``cleaned'' versions of RVL-CDIP for benchmarking.
These new versions of RVL-CDIP are corrected and cleaned, and minimize test-train duplication.
With these new clean versions of RVL-CDIP we are then able to study the impact of label errors and test-train overlap on several document classification models, including transformer-based and zero-shot models.
We find that model performance increases upon removing label errors, but decreases when duplicates are removed.
We additionally evaluate our models on RVL-CDIP-N~\cite{larson-etal-2022-rvlcdipn}, an out-of-distribution (OOD) benchmark for RVL-CDIP, to assess whether training on cleaner data improves generalization beyond RVL-CDIP's tobacco-industry domain.
We make our cleaned RVL-CDIP versions publicly available to aid in benchmarking document classification models on cleaner, more reliable data.\footnote{\url{https://rvlcdip-errors.com/}}

\begin{table}
\centering
%todo: verify email (24,989) and scientific_publication (24,999) counts -- all categories should be 25,000
\caption{Dataset statistics for RVL-CDIP. Train, validation, and test split sizes per category.}
\label{tab:rvlcdip-statistics}
\scalebox{0.85}{
\begin{tabular}{lccc}
\toprule
\textbf{Category} & \textbf{Train} & \textbf{Val.} & \textbf{Test} \\
\midrule
\texttt{advertisement} & 19,963 & 2,522 & 2,515 \\
\texttt{budget} & 20,010 & 2,485 & 2,505 \\
\texttt{email} & 19,954 & 2,530 & 2,505 \\
\texttt{file\_folder} & 20,022 & 2,451 & 2,527 \\
\texttt{form} & 19,957 & 2,537 & 2,506 \\
\texttt{handwritten} & 20,034 & 2,434 & 2,532 \\
\texttt{invoice} & 19,947 & 2,576 & 2,477 \\
\texttt{letter} & 20,106 & 2,430 & 2,464 \\
\texttt{memo} & 19,975 & 2,533 & 2,492 \\
\texttt{news\_article} & 20,011 & 2,526 & 2,463 \\
\texttt{presentation} & 20,043 & 2,468 & 2,489 \\
\texttt{questionnaire} & 20,048 & 2,517 & 2,435 \\
\texttt{resume} & 20,037 & 2,426 & 2,537 \\
\texttt{scientific\_publication} & 19,902 & 2,526 & 2,571 \\
\texttt{scientific\_report} & 19,994 & 2,508 & 2,498 \\
\texttt{specification} & 19,997 & 2,531 & 2,472 \\
\bottomrule
\end{tabular}}
\end{table}

This paper is a \emph{benchmark audit}: we propose no new model or training procedure.
Instead, our contribution is to provide the first exhaustive, empirically grounded analysis of data quality in the document classification community's primary benchmark, and to quantify the concrete impact of these issues on reported model performance. This type of paper is in the tradition of prior work that has revealed similar problems across NLP, computer vision, and document understanding benchmarks (see Section~\ref{sec:Background}).

To summarize, the main contributions of this paper are:
\begin{enumerate}
    \item We conduct the first exhaustive, 400,000-document review of RVL-CDIP, finding that approximately 12\% of the dataset is erroneously labeled --- a finding with implications for every performance claim made against this benchmark.
    \item We develop a method to detect test-train overlap, finding roughly 35\% of RVL-CDIP's test set to have a (near-) duplicate in the training set.
    \item We create cleaned versions of RVL-CDIP free of label error and test-train duplication and re-benchmark several document classification models.
    \item We evaluate our models on RVL-CDIP-N, an out-of-distribution benchmark, to assess the effect of training data quality on OOD generalization.
\end{enumerate}

\section{Background and Related Work}\label{sec:Background}

RVL-CDIP is widely used to benchmark document page classifiers.
Larson et al.~\cite{larson-etal-2023-evaluation} called it the \emph{de facto} standard benchmark for document page classification, and when we surveyed 30 papers that used RVL-CDIP to benchmark document classification systems or models, roughly 95\% \emph{exclusively} used RVL-CDIP (or a subset of RVL-CDIP) for performance evaluation.
A wide range of model types have used RVL-CDIP to benchmark classification performance, including image-based models, text-based models, multi-modal models, zero-shot models, and Large Language Models.
RVL-CDIP is therefore not merely the most popular document classification benchmark --- it is effectively the \emph{only} standardized benchmark of comparable scale for this task, making a rigorous audit of its quality a critical service to the document understanding community.

The standard benchmarking setup with RVL-CDIP is 16-class classification.
RVL-CDIP consists of 400,000 samples spread roughly evenly across its 16 categories.
These categories and their counts are listed in Table~\ref{tab:rvlcdip-statistics}; samples of RVL-CDIP are displayed in Figure~\ref{fig:rvlcdip-examples}.
Harley et al.~\cite{harley2015icdar-rvlcdip}, who introduced RVL-CDIP, provided no definitions or distinguishing criteria for the 16 categories, but noted that the source collection had ``missing or erroneous tags,'' and acknowledged that ``the final categories are not perfectly distinct.''
This absence of official category definitions means the research community has been benchmarking against a dataset with no authoritative criteria for what each category contains --- a gap that our annotation work in Section~\ref{sec:error-detection} directly addresses.

In the supervised learning setting, models are trained on RVL-CDIP's training set and evaluated on the test set to establish an accuracy score.
In zero-shot settings (e.g., with LLMs or zero-shot image classifiers), models are evaluated on RVL-CDIP's test set only.
To our knowledge, the state-of-the-art model is LayoutLLM, with a reported accuracy of 98.8\%~\cite{fujitake-2024-layoutllm}, followed closely by EAML (97.70\%; \cite{eaml-bakkali-2021}) and Bi-VLDoc (97.17\%; \cite{bi-vldoc-2025-luo}).

\begin{table}
    \centering
    \caption{Label error breakdown in RVL-CDIP. In total, we determine that RVL-CDIP contains approximately 12\% label error.}
    \label{tab:rvl_cdip_errors}
    \scalebox{0.62}{
    \begin{tabular}{lcccccc}
        \toprule
        \textbf{Category} & \textbf{Unknown} & \textbf{Wrong} & \textbf{Mixed} & \textbf{Any Error} & \textbf{Unsure} & \textbf{Any + Unsure} \\
        \midrule
        \texttt{advertisement} & $7.51\%$ & $0.60\%$ & $1.62\%$ & $9.74\%$ & $1.50\%$ & $11.24\%$ \\
        \texttt{budget} & $8.87\%$ & $8.48\%$ & $3.94\%$ & $21.29\%$ & $1.66\%$ & $22.94\%$ \\
        \texttt{email} & $8.24\%$ & $1.27\%$ & $0.12\%$ & $9.63\%$ & $0.50\%$ & $10.13\%$ \\
        \texttt{file\_folder} & $3.85\%$ & $0.16\%$ & $0.52\%$ & $4.53\%$ & $0.08\%$ & $4.61\%$ \\
        \texttt{form} & $10.63\%$ & $3.62\%$ & $6.10\%$ & $20.35\%$ & $1.04\%$ & $21.39\%$ \\
        \texttt{handwritten} & $10.51\%$ & $2.14\%$ & $1.03\%$ & $13.68\%$ & $0.00\%$ & $13.69\%$ \\
        \texttt{invoice} & $6.42\%$ & $8.52\%$ & $0.59\%$ & $15.53\%$ & $1.16\%$ & $16.70\%$ \\
        \texttt{letter} & $15.75\%$ & $15.51\%$ & $0.42\%$ & $31.68\%$ & $0.04\%$ & $31.72\%$ \\
        \texttt{memo} & $4.23\%$ & $1.72\%$ & $1.12\%$ & $7.08\%$ & $1.91\%$ & $8.99\%$ \\
        \texttt{news\_article} & $4.05\%$ & $4.50\%$ & $0.54\%$ & $9.09\%$ & $0.96\%$ & $10.05\%$ \\
        \texttt{presentation} & $4.13\%$ & $1.34\%$ & $0.87\%$ & $6.34\%$ & $2.08\%$ & $8.42\%$ \\
        \texttt{questionnaire} & $8.47\%$ & $5.98\%$ & $1.09\%$ & $15.54\%$ & $1.27\%$ & $16.81\%$ \\
        \texttt{resume} & $1.48\%$ & $0.01\%$ & $0.00\%$ & $1.49\%$ & $0.01\%$ & $1.50\%$ \\
        \texttt{scientific\_publication} & $2.15\%$ & $2.39\%$ & $0.00\%$ & $4.54\%$ & $2.72\%$ & $7.26\%$ \\
        \texttt{scientific\_report} & $8.07\%$ & $4.09\%$ & $5.17\%$ & $17.32\%$ & $0.45\%$ & $17.77\%$ \\
        \texttt{specification} & $2.29\%$ & $1.13\%$ & $0.74\%$ & $4.16\%$ & $0.24\%$ & $4.40\%$ \\
        \midrule
        RVL-CDIP & $6.67\%$ & $3.84\%$ & $1.49\%$ & \textbf{$12.00\%$} & $0.98\%$ & $12.98\%$ \\
        \bottomrule
    \end{tabular}}
\end{table}

Recent work by Larson et al.~\cite{larson-etal-2023-evaluation} raised several issues with RVL-CDIP, including the presence of label errors and data duplication in the dataset.
Given the role that RVL-CDIP plays in the document understanding research area as the \emph{de facto} benchmark for classification, these data quality issues are alarming, as they call into question the validity of certain claims like state-of-the-art performance.
In our present work, we aim to rigorously measure the extent of label errors and test-train overlap in RVL-CDIP, and to quantify their impact on model performance.
In this way, our paper fits within relevant literature on analyzing data quality in benchmark datasets.
This prior work includes investigating data label issues in text (e.g.,~\cite{niu-penn-2019-rationally-atis,wang-conll-2019-errors,croft-2023-data-quality-vul,rucker-akbik-2023-cleanconll}), image (e.g.,~\cite{radenovic-2018-revisiting-oxford-paris,muller-ijcnn-mislabeled-2019,pervasive-label-errors-2021,li-2022-automotive-errors,cathaoir-2023-coco-data-quality,pmlr-v225-groger23a}), and document~\cite{vu-2020-funsd-quality,larson-etal-2023-evaluation,scius-bertrand-2023-handwriting-data,lim-2025-tobacco3482-quality} datasets,
along with investigating data duplication and test-train overlap issues in text (e.g.,~\cite{allamanis-2019-adverse-duplication,lewis-etal-2021-question,mu-etal-2024-enhancing}), image (e.g.,~\cite{cifar-duplicates-2020,laroca-overlap-license-plate-ijcnn-2023,face-overlap-2024-hu}), and document (e.g.,~\cite{laatiri-2023-funsd-sroie-overlap,larson-etal-2023-evaluation}) benchmarks.
Collectively, these works demonstrate that label errors and test-train overlap are pervasive problems across benchmark datasets, and that addressing them often changes the interpretation of reported model performance.

In particular, Larson et al.~\cite{larson-etal-2023-evaluation} estimated that 9.7\% of RVL-CDIP has label errors, and 32\% of the benchmark's test set has a duplicate or near-duplicate in the train set.
Our present work goes beyond the approximation of Larson et al.~\cite{larson-etal-2023-evaluation} and more rigorously quantifies the amount of label error and test-train overlap present in RVL-CDIP.
We create cleaned versions of RVL-CDIP with errors and duplicates removed and compute model performance on this cleaned data, which was not explored by Larson et al.~\cite{larson-etal-2023-evaluation}.
Separately, Larson et al.~\cite{larson-etal-2022-rvlcdipn} introduced RVL-CDIP-N, an out-of-distribution benchmark for RVL-CDIP, finding that models trained on RVL-CDIP do not generalize well beyond its tobacco-industry domain; we include RVL-CDIP-N in our evaluation.

\section{Label Error Detection}\label{sec:error-detection}
Unlike prior work that estimated label error rates through sampling~\cite{larson-etal-2023-evaluation}, we exhaustively review all 400,000 documents in RVL-CDIP for label errors.

\begin{figure*}
    \centering
    \includegraphics[width=\linewidth]{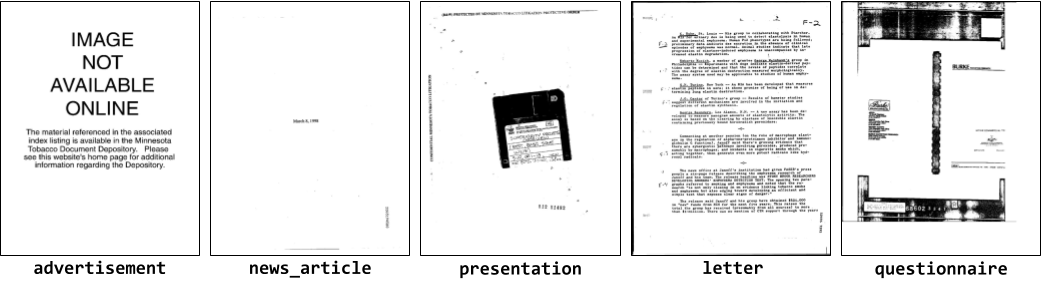}
    \caption{Examples of RVL-CDIP label errors with no valid true label.}
    \label{fig:unknown_errors}
\end{figure*}

\begin{figure*}
    \centering
    \includegraphics[width=\linewidth]{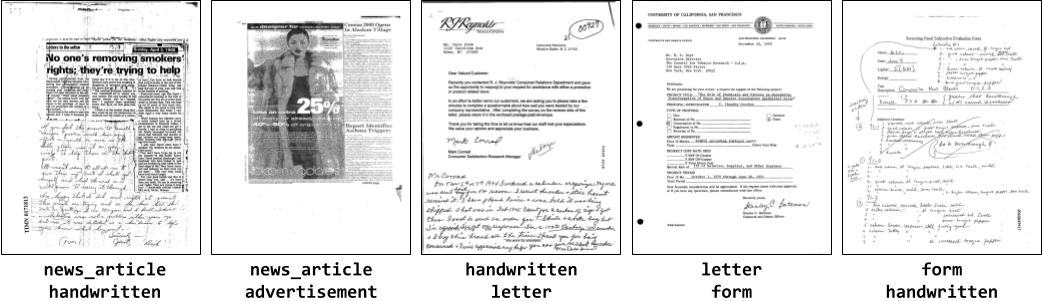}
    \caption{Examples from RVL-CDIP with multiple valid labels. Top labels: original; bottom labels: alternate label.}
    \label{fig:mixed-examples}
\end{figure*}

\subsection{Annotation}

Our goal is to measure the amount of label error in RVL-CDIP.
Possible tools to help accomplish this goal include \texttt{CleanLab}\footnote{\url{https://github.com/cleanlab/cleanlab}}, which uses confident learning~\cite{pervasive-label-errors-2021}, but this approach is not feasible here because RVL-CDIP's pervasive and systematic label errors corrupt the model confidence scores that confident learning relies on --- a conclusion also reached by Larson et al.~\cite{larson-etal-2023-evaluation} in their preliminary investigation.
Instead, we manually inspect all documents from RVL-CDIP using the label descriptions and criteria outlined by Larson et al.~\cite{larson-etal-2023-evaluation}.
We re-print these criteria on our companion website.\footnote{\url{https://rvlcdip-errors.com/}}
There are three error types that we annotate in RVL-CDIP: (1) \emph{unknown}, where we determine that the document does not have a valid RVL-CDIP label; (2) \emph{wrong label}, where we determine the document has a wrong original label but a valid, more-correct label; (3) \emph{mixed}, where we determine that the document has multiple valid labels, including the original.

A team of seven volunteer undergraduate researchers annotated RVL-CDIP over the span of roughly two years.
The annotators were given document images in batches grouped by category along with descriptions of each category.
Each batch contained 1,000 samples from one particular category.
The annotators were instructed to provide one of the following annotations: ``yes'' if the document fit the provided category criteria,
``no'' if the document clearly did not fit the category criteria, or ``unsure'' in cases of ambiguity.
The project lead reviewed all ``no'' and ``unsure'' cases as well as a portion of ``yes'' cases in order to provide corrective instruction and to update the category criteria if necessary.
Documents marked as ``no'' were also given a corrected label, where appropriate.
Thus this annotation process was iterative, with category criteria refined continuously as annotators encountered ambiguous cases.
In some cases, documents might fit within multiple category criteria, in which the document was marked as ``mixed'' and the multiple categories to which that document belonged were also recorded.
Inter-rater agreement was measured on a sample of 1,500 documents across 5 categories (\texttt{resume}, \texttt{letter}, \texttt{memo}, \texttt{handwritten}, and \texttt{email}), which yielded a Fleiss's kappa score of 0.74, indicating substantial agreement~\cite{landis-koch-1977}.

\begin{figure*}
    \centering
    \includegraphics[width=\linewidth]{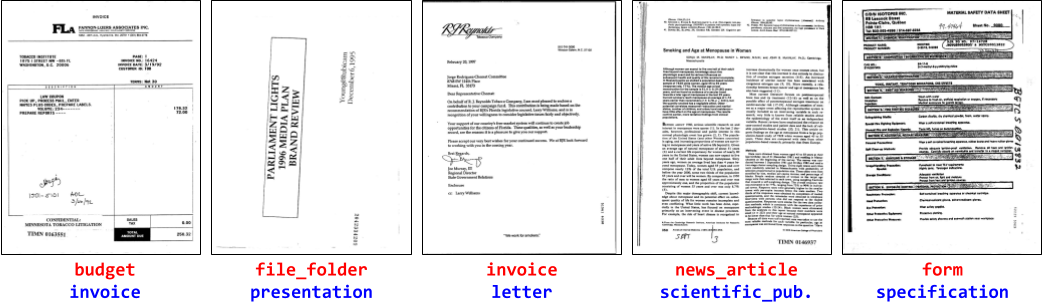}
    \caption{Examples of incorrectly labeled samples from RVL-CDIP with a known correct label.}
    \label{fig:wrong_labels_examples}
\end{figure*}

\subsection{Findings}

Through our exhaustive annotation process, we find that RVL-CDIP contains many label errors.
Table~\ref{tab:rvl_cdip_errors} lists a per-category breakdown of these errors.
Categories with the highest error rates include \texttt{letter}, with roughly 31.7\% of that category being erroneously labeled, but \texttt{budget}, \texttt{form}, \texttt{handwritten}, \texttt{invoice}, \texttt{questionnaire}, and \texttt{scientific\_report} also each have more than 10\% label error.
Examples of various error types are shown in Figure~\ref{fig:unknown_errors} (\emph{unknown} errors), Figure~\ref{fig:wrong_labels_examples} (\emph{wrong} errors), and Figure~\ref{fig:mixed-examples} (\emph{mixed} errors).
We note that the 12\% label error rate is higher than the rough estimate of Larson et al.~\cite{larson-etal-2023-evaluation}, who reported 9.7\% (they reported 8.1\% for \emph{wrong} and \emph{unknown}, and 1.7\% for \emph{mixed}).
Our approach is more exhaustive and thorough than that of Larson et al.~\cite{larson-etal-2023-evaluation}, who relied on sampling.
The exhaustive nature of our review meant that annotators developed a deeper understanding of category boundaries over time, and were better able to identify subtle errors that sampling-based approaches may miss --- particularly systematic error patterns that only become apparent at scale.
For example, press release documents appear in both the \texttt{news\_article} and \texttt{presentation} categories, but are more prevalent in \texttt{presentation}; this pattern was only recognizable after reviewing documents at scale.
%todo: find example press release documents mislabeled as news\_article vs presentation to show

\begin{framed}
    \noindent
    \textbf{Main finding:} RVL-CDIP contains approximately 12\% label error. The dataset's \texttt{letter} category contains the highest number of errors (31.68\%).
\end{framed}

\section{Quantifying Test-Train Overlap}\label{sec:duplication}

Our goal is to quantify the amount of overlap between the official test and train splits of RVL-CDIP, and if a non-trivial amount of overlap is found, to re-compute model performance scores on a modified version of the test set where duplicates and near-duplicates are removed.
Prior work has found non-trivial amounts of overlap between the test and train splits of other datasets (see Section~\ref{sec:Background}); in the words of Allamanis~\cite{allamanis-2019-adverse-duplication}, large amounts of data duplication across test and train splits can lead to ``inflated'' performance scores because a portion of the test set has been ``seen'' by the model during training.
Thus it is important to ensure datasets minimize the amount of test-train overlap in order to facilitate meaningful model benchmarking.

\begin{figure}
    \centering
    \includegraphics[width=0.87\linewidth]{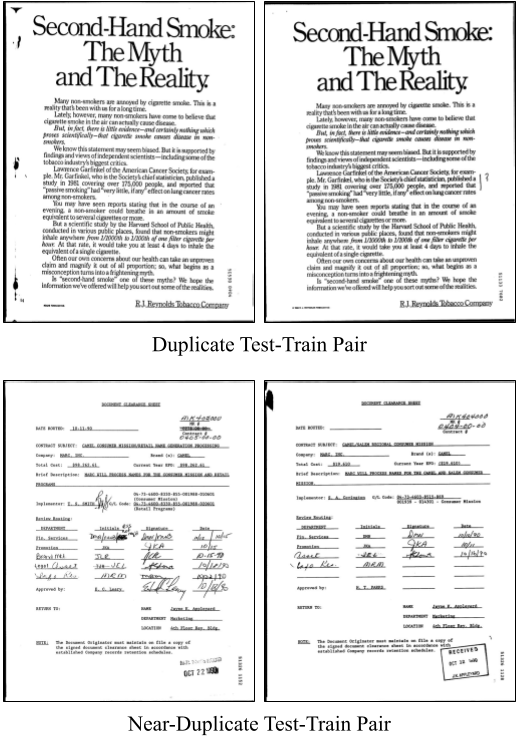}
    \caption{Example duplicate (top) and near-duplicate (bottom) test-train pairs. Near-duplicates are not pixel-identical: they share the same underlying document template but may differ in filled fields or scan conditions. Both types constitute test-train leakage.}
    \label{fig:duplicate-types}
\end{figure}

\subsection{Data Similarity Analysis Methods}\label{subsec:duplication-methods}
We seek samples in RVL-CDIP's test set that have a duplicate or near-duplicate in the train set.
We consider a pair of two documents to be (near-) duplicates if they are
(1) nearly exact copies of the same document, with one of the pair perhaps containing different amounts of noise or geometric distortion than the other;
(2) a document template match, meaning both documents in the pair are slightly different manifestations of the same template document.
This template can be a form that is filled out slightly differently, or the same letter that is addressed to two different addressees but bears the same content.
Examples of cases (1) and (2) are shown in Figure~\ref{fig:duplicate-types}, where they are called duplicate and near-duplicate pairs, respectively.
While we do not distinguish between these two types in the remainder of this paper, we point them out here because they help inform us of the space of data similarity analysis methods to consider.
Importantly, both types constitute meaningful test-train leakage: a model trained on a near-duplicate of a test document has effectively been exposed to that test instance during training, regardless of whether the two documents are pixel-identical copies.

Since we consider two documents to be (near-) duplicates even if their pixels differ significantly (e.g., due to scanner noise or geometric transformation), image hashing algorithms like pHash\footnote{\url{https://www.phash.org/}}---used in open source tools like CleanVision\footnote{\url{https://github.com/cleanlab/cleanvision}} and fastdup\footnote{\url{https://github.com/visual-layer/fastdup}}---are not applicable, as they will suffer from low recall.
Instead, we consider a filter-and-refine pipeline that uses approximate similarity to propose candidate match pairs.
Then, more exact matching methods are used to verify if a candidate match pair is indeed a (near-) duplicate pair.

The ``filter'' step of this process involves computing rough similarity scores between documents using page embeddings.
Similarity scores can be efficiently computed between the test set and the training set with matrix multiplication: if we stack $d$-dimensional $\ell_2$-scaled embedding vectors of data from train and test sets into matrices $X_{train}$ and $X_{test}$, then the resulting matrix of cosine similarity scores is $S = X_{train}X_{test}^{\top}$.
In this work, we use CLIP~\cite{clip-2021} image embeddings ($d=512$).
For each test set sample, we find the 10 most similar training set documents using this approach.
We then pass those 10 candidate match documents on to the ``refine'' step where they are more thoroughly compared against the test sample.

In the ``refine'' step, we aim to verify if candidate test-train pairs selected based on the ``filter'' step are actually duplicates.
We consider two approaches: (1) image feature matching, where local features within two images are compared, and (2) text sequence similarity.
The image feature matching method we investigate uses SuperPoint~\cite{superpoint-2018} for feature extraction and LightGlue~\cite{lindenberger2023lightglue} for feature matching.
SuperPoint is a self-supervised convolutional neural network that detects keypoints and computes descriptors from image patches.
LightGlue is a lightweight graph neural network that matches keypoints between two images based on these descriptors.
This pairing is well-suited for detecting near-duplicates in RVL-CDIP, as SuperPoint was designed to handle geometric transformations such as the distortion that can differentiate (near-) duplicate document scans.
The text sequence similarity method we investigate uses the Ratcliff/Obershelp pattern recognition algorithm~\cite{difflib-gestalt}, implemented via Python's \texttt{difflib} library\footnote{\url{https://docs.python.org/3/library/difflib.html}}, which recursively finds the longest common substring between two sequences to compute an overall similarity ratio.
This makes it well-suited for detecting template-match near-duplicates, where two documents share a large body of common text but differ in small regions (e.g., a filled-in field or a different addressee).
In practice, we use a decision rule where we consider two documents to be a (near-) duplicate match if the sequence similarity is $\ge 0.2$ and the feature match score (i.e., the number of features matched between the two documents) is $\ge 100$.
These thresholds were selected based on tuning on a ground truth dataset of 1,600 document pairs (100 test samples from each of the 16 categories), where candidate pairs were generated using the CLIP-based filter step and then manually reviewed by the authors.
This ground-truth dataset consists of 403 match pairs.
All ground truth pairs are available on our companion website.\footnote{\url{https://rvlcdip-errors.com/}}
The decision rule settings that we use achieve a precision of 100\%, recall of 82.9\%, and F1 score of 90.6\% on this calibration dataset.
%todo C: add class balance (# actual duplicates vs. non-duplicates out of 1,600)
%todo D: add how thresholds were selected (grid search? ROC curve?)
%todo A: consider a Prolific crowdworker study to validate ground truth reliability

\subsection{Findings}
\begin{figure*}
    \centering
    \scalebox{0.335}{
    \includegraphics{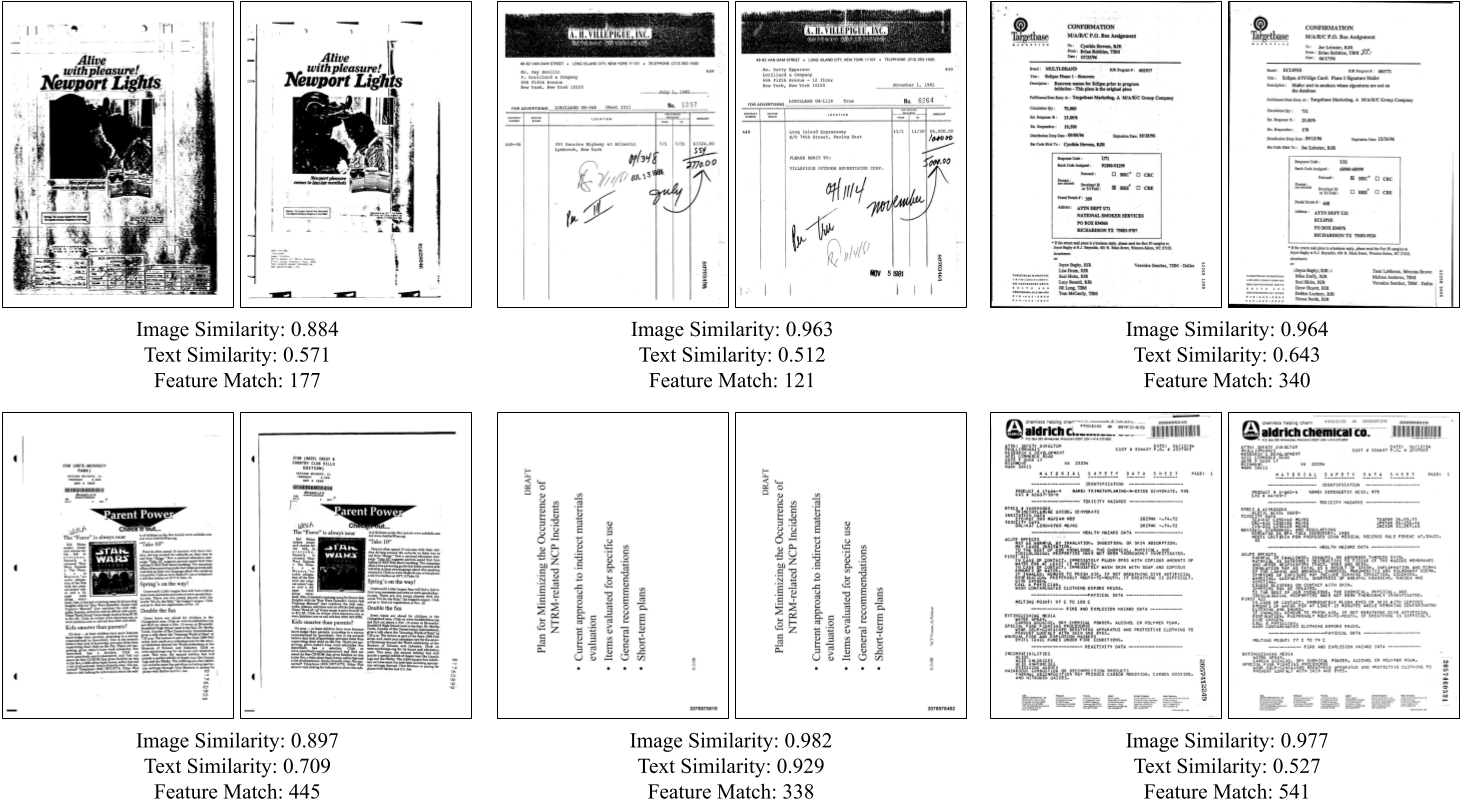}}
    \caption{Example (near-) duplicate test-train pairs. Documents need not be pixel-identical to constitute leakage; template-match pairs share the same underlying document structure, effectively exposing the model to the test document during training.}
    \label{fig:test-train-pairs}
\end{figure*}

We run our duplicate detection method on RVL-CDIP, and find that a large portion of RVL-CDIP's test set has a (near-) duplicate training sample.
Examples of (near-) duplicates are shown in Figure~\ref{fig:test-train-pairs}.
A breakdown of the number and percentage of (near-) duplicates for each RVL-CDIP category is shown in Table~\ref{tab:rvl_cdip_dups}.
The more extreme cases are
\texttt{budget}, \texttt{form}, \texttt{invoice}, \texttt{questionnaire}, \texttt{resume}, and \texttt{specification}, each of which has more than half of their test sets having a duplicate (or near-duplicate) document in the training set.
Half (8 out of 16) RVL-CDIP categories have over one-third of their test sets having a corresponding duplicate in the train set (see Table~\ref{tab:rvl_cdip_dups}).
On average, approximately 35\% of RVL-CDIP's test set has a duplicate or near-duplicate counterpart in the training set.
We additionally apply our method to RVL-CDIP's validation set, finding a nearly identical rate: approximately 35\% of the validation set also has a (near-) duplicate in the training set (Table~\ref{tab:rvl_cdip_dups}).
The close agreement between test-train and val-train duplication rates suggests that duplication is a systematic property of how RVL-CDIP was constructed, rather than an artifact of any particular split.

\begin{table}
\centering
\caption{Test-train and val-train duplication by RVL-CDIP category. The \% Dup. percentage indicates the portion of the split that contains a (near-) duplicate in the training set.}
\label{tab:rvl_cdip_dups}
\scalebox{0.8}{
\begin{tabular}{lcccc}
\toprule
\textbf{Category} & \multicolumn{2}{c}{\textbf{Test}} & \multicolumn{2}{c}{\textbf{Val.}} \\
\cmidrule(lr){2-3}\cmidrule(lr){4-5}
& \textbf{\# Dup.} & \textbf{\% Dup.} & \textbf{\# Dup.} & \textbf{\% Dup.} \\
\midrule
\texttt{advertisement} & 856 & 34.04\% & 843 & 33.43\% \\
\texttt{budget} & 1,479 & 59.04\% & 1,450 & 58.35\% \\
\texttt{email} & 175 & 6.99\% & 190 & 7.51\% \\
\texttt{file\_folder} & 50 & 1.98\% & 56 & 2.28\% \\
\texttt{form} & 1,458 & 58.18\% & 1,455 & 57.35\% \\
\texttt{handwritten} & 102 & 4.03\% & 112 & 4.60\% \\
\texttt{invoice} & 1,703 & 68.75\% & 1,803 & 70.00\% \\
\texttt{letter} & 292 & 11.85\% & 259 & 10.66\% \\
\texttt{memo} & 457 & 18.34\% & 440 & 17.37\% \\
\texttt{news\_article} & 501 & 20.34\% & 514 & 20.35\% \\
\texttt{presentation} & 710 & 28.53\% & 688 & 27.88\% \\
\texttt{questionnaire} & 1,499 & 61.56\% & 1,515 & 60.19\% \\
\texttt{resume} & 1,461 & 57.59\% & 1,360 & 56.06\% \\
\texttt{scientific\_publication} & 445 & 17.30\% & 450 & 17.81\% \\
\texttt{scientific\_report} & 856 & 34.27\% & 899 & 35.84\% \\
\texttt{specification} & 2,032 & 82.20\% & 2,117 & 83.64\% \\
\midrule
Total & 14,076 & 35.20\% & 14,151 & 35.38\% \\
\bottomrule
\end{tabular}}
\end{table}

\begin{framed}
    \noindent
    \textbf{Main finding:} RVL-CDIP's test set has many samples that are highly similar to the training set. More than one-third ($\sim 35\%$) of the test set has a duplicate or near-duplicate sample in the training set, including template-match near-duplicates where shared document structure constitutes effective data leakage even without pixel-identical copies.
\end{framed}

\section{Evaluation on Cleaned RVL-CDIP Versions}\label{sec:new-versions}

Now that we have rigorously identified label issues and duplication issues, we seek to investigate the impact of these issues on document classification performance.
We do this by creating different versions of RVL-CDIP with various data quality issues addressed.
The rest of this section is as follows: Section~\ref{subsec:models} introduces the models that we benchmark, Section~\ref{subsec:rvlcdip-variants} discusses our modifications to RVL-CDIP, and Section~\ref{subsec:rvlcdip-n} discusses RVL-CDIP-N's OOD data.

\subsection{Models}\label{subsec:models}

We investigate the impact of data label issues on several classes of document classification models.
The supervised learning models we investigate include text-only, image-only, and multi-modal models.
For the text-only modality, we investigate BERT~\cite{devlin-etal-2019-bert}, RoBERTa~\cite{roberta}, LongFormer~\cite{Beltagy2020Longformer}, and Big Bird~\cite{zaheer2020bigbird}.
For the image-only modality, we investigate the CNNs VGG-16~\cite{vgg-16}, ResNet-50~\cite{res-net-50}, ResNeXT-50~\cite{resnext-Xie2016}, GoogLeNet~\cite{googlenet},  AlexNet~\cite{alexnet}, DocXclassifier~\cite{Saifullah2024-docxclassifier}, as well as DiT~\cite{li-2022-dit-image-transformer} and Donut~\cite{kim2022donut}.
The supervised multi-modal models we use are LayoutLM~\cite{xu-2020-layoutlm} and LayoutLMv3~\cite{layoutlmv3-huang-2022}.

Prior work has benchmarked pre-trained Large Language Models (LLMs) in a zero-shot setting on RVL-CDIP as well~\cite{llms-rvlcdip-zero-shot-2024}, so we also include
these in our investigation.
We experiment with gpt-oss-20b~\cite{openai2025gptoss120bgptoss20bmodel}, Llama 4 Maverick 17B~\cite{meta2025llama4}, Qwen 3 32b \cite{yang2025qwen3technicalreport}, and Mistral 7B Instruct~\cite{mistral-7b}.
We additionally evaluate vision-language models in a zero-shot setting, including CLIP~\cite{clip-2021}, ALIGN~\cite{align}, SigLIP~\cite{siglip-2023}, and SigLIP~2~\cite{tschannen2025siglip}.\footnote{We use the Hugging Face implementations \texttt{openai/clip-vit-large-patch14-336} for CLIP, \texttt{kakaobrain/align-base} for ALIGN, \texttt{google/siglip-so400m-patch14-384} for SigLIP, and \texttt{google/siglip2-so400m-patch16-512} for SigLIP~2.}
We measure document classification accuracy for all models.
Experimental settings such as libraries and hyperparameters used are available on our companion website.
We do not re-train Donut or DiT due to resource constraints, and use a pre-trained version of these supervised models only.

\subsection{RVL-CDIP Variants}\label{subsec:rvlcdip-variants}
We use the following versions of RVL-CDIP in our benchmarking (label errors and duplicates are detected using the approaches discussed in Sections~\ref{sec:error-detection} and~\ref{sec:duplication}):

\begin{enumerate}
\itemsep0em
    \item \textbf{Original RVL-CDIP}: the original RVL-CDIP from Harley et al.~\cite{harley2015icdar-rvlcdip}. We train models on this version's training set.
    \item \textbf{Errors Dropped}: where all erroneously labeled test samples are removed.
    \item \textbf{Errors Fixed}: where all erroneously labeled test samples are fixed. This means that \emph{unknown} and \emph{mixed} documents are removed, and \emph{wrong} label documents have their labels fixed (\emph{unsure} documents are also removed).
    % TODO would be interesting if we included mixed but allowed for multiple valid labels when computing accuracy
    \item \textbf{Fixed + Retrain}: where all erroneously labeled samples in RVL-CDIP are fixed; we retrain models on this version's training set.
    \item \textbf{Original De-Duplicated}: where test samples flagged as having a (near-) duplicate counterpart in the train set are removed.
    \item \textbf{Errors Dropped and De-Duplicated}: where duplicate test samples are dropped (if they have a duplicate train counterpart) and all errors are dropped from RVL-CDIP's test set.
    \item \textbf{Errors Fixed and De-Duplicated}: where duplicate test samples are dropped (if they have a duplicate train counterpart) and all erroneously labeled test samples are fixed.
    \item \textbf{Fixed + Retrain and De-Duplicated}: where we fix all errors in all of RVL-CDIP, and remove test duplicates if they have a (near-) duplicate train pair.
\end{enumerate}

%\begin{figure}
%    \centering
%    \includegraphics[width=0.95\linewidth]{figures/rvlcdip_n_samples.png}
%    \caption{Examples from RVL-CDIP-N \cite{larson-etal-2022-rvlcdipn}.}
%    \label{fig:rvl_cdip_n}
%\end{figure}

% CLIP: ViT-L_14@336px
% SigLIP: siglip-so400m-patch14-384
% SigLIP2: siglip2-so400m-patch16-512
\begin{table*}[h]
\centering
\caption{Model performance on RVL-CDIP with different data configurations. ``De-Dup.'' means test samples that were determined to have duplicates in the train set were dropped (see Section~\ref{sec:duplication}). Overall, model accuracy scores increase when data label issues are addressed by removing those test samples or re-training on error-free data. Accuracy scores decrease when duplicates are removed from the test set.}
\label{tab:model_performance}
\scalebox{0.85}{
\begin{tabular}{lcccccccccc}
\toprule
 & & \textbf{Original} & \textbf{Original} & \textbf{Errors} & \textbf{Errors} & \textbf{Fixed +} & \textbf{Original} & \textbf{Err. Dr.} & \textbf{Err. Fix.} & \textbf{Fix. Ret.} \\
\textbf{Model} & \textbf{Modality} & \textbf{Reported} & \textbf{Ours} & \textbf{Dropped} & \textbf{Fixed} & \textbf{Retrain} & \textbf{De-Dup.} & \textbf{De-Dup.} &\textbf{De-Dup.} & \textbf{De-Dup.} \\
\midrule
BERT & Text & 89.81 & 92.44 & 95.53 & 93.27 & 95.94 &89.47 &93.59 &91.16 & 94.22 \\
RoBERTa & Text & 90.06 & 92.71 & 95.73 & 93.44 & 96.14 & 89.87 & 93.91 & 91.48 & 94.56 \\
LongFormer & Text & 93.85 & 93.11 & 96.03 & 93.75 & 96.11 & 90.62 & 94.43 & 92.95 & 94.65 \\
Big Bird & Text & 93.48 & 92.10 & 95.30 & 93.08 & 95.71 & 89.20 & 93.38 & 91.99 & 94.05 \\
\hline
VGG-16 & Image & 90.97 & 90.99 & 94.01 & 91.18 & 94.26 & 87.73 & 91.70 & 89.31 & 92.11 \\
ResNet-50 & Image & 90.40 & 89.88 & 93.19 & 90.42 & 93.38 & 86.70 & 90.95 & 88.58& 91.36 \\
ResNeXT-50 & Image & --- & 90.48 & 93.71 & 90.94 & 93.90 & 87.28 &91.54 &89.17 & 91.90\\
GoogLeNet & Image & 89.02 & 87.78 & 91.26 & 88.56 & 91.43 & 84.48 &  89.00& 86.64& 89.37 \\
AlexNet & Image & 88.60 & 88.07 & 91.41 & 88.48 & 91.93 & 84.51 & 88.81 & 86.50 & 89.42 \\
DocXclassifier & Image & 94.00 & 94.04 & 96.71 & 94.14 & --- & 91.71 & 95.29 & 93.45 & --- \\
DiT & Image & 92.11 & 93.28 & 96.26 & 93.76 & --- & 90.84 & 94.75 & 92.94 & --- \\
Donut & Image & 95.30 & 95.24 & 97.57 & 94.88 & --- &93.44 &96.58 & 94.06& --- \\
\hline
LayoutLM & Multi & 94.42 & 93.48 & 96.47 & 94.18 & 94.35 & 91.11 & 95.03 & 93.54 & 93.79 \\
LayoutLMv3 & Multi & 95.44 & 94.30 & 96.87 & 94.24 & 97.32 &92.34 &95.73 &93.81 & 96.36 \\
\hline
CLIP & Image & --- & 42.26 & 45.54 & 45.32 & --- & 42.69 & 47.01 & 46.87 & --- \\
ALIGN & Image & --- & 46.95 & 50.98 & 51.14 & --- & 47.07 & 52.02 & 52.15 & --- \\
SigLIP & Image & --- & 56.69 & 61.90 & 61.66 & --- & 56.69 & 61.71 & 62.84 & --- \\
SigLIP 2 & Image & --- & 57.93 & 62.89 & 62.57 & --- & 56.24 & 61.55 & 61.45 & --- \\
\hline
gpt-oss-20b & Text & --- & 65.44 & 71.08 & 70.44 & --- & 64.47 & 70.89 &69.05 & --- \\
%gpt-oss-120b & Text & --- & 66.87 & 72.68 & 72.05 & --- & 65.30 & 71.91 & 70.04 & --- \\
Mistral 7B Instruct & Text & --- & 56.96 & 61.56 & 60.86 & --- & 54.35 & 59.75 & 59.43 & --- \\
Llama 4 Maverick 17B & Text & --- & 67.61 & 73.47 & 72.78 & --- & 65.28 & 72.03 & 71.55 & --- \\
Qwen 3 32b & Text & --- & 62.28 & 67.27 & 66.38 & --- & 62.61 & 68.92 & 68.47 & --- \\
\bottomrule
\end{tabular}}
\end{table*}

\subsection{Out-of-Distribution Data}\label{subsec:rvlcdip-n}
To evaluate generalization beyond RVL-CDIP's tobacco-industry domain, Larson et al.~\cite{larson-etal-2022-rvlcdipn} created RVL-CDIP-N, an out-of-distribution benchmark of 1,002 documents that span the same 16 categories as RVL-CDIP but are sourced from DocumentCloud\footnote{\url{https://www.documentcloud.org/home/}} and web search rather than the Legacy Tobacco Document Library.
Examples from RVL-CDIP-N are available on our companion website.
Larson et al.\ found that certain models trained on RVL-CDIP exhibit accuracy drops of roughly 15--30\% on RVL-CDIP-N, suggesting that high in-distribution accuracy does not reflect general classification ability.
We evaluate our selected models on RVL-CDIP-N to check for performance differences across versions of RVL-CDIP training data---in particular, whether training on error-corrected data (Fixed + Retrain) improves OOD generalization.

\section{Results}\label{sec:results}

\subsection{In-Distribution Results}

Document classification accuracy scores for each model are listed in Table~\ref{tab:model_performance}. Before comparing across conditions, we note that our replicated Original scores are generally consistent with published results. Text-based models BERT and RoBERTa are exceptions, scoring +2.63 percentage points (pp) and +2.65~pp above their reported scores respectively; we attribute this to our use of AWS Textract as the OCR engine rather than Tesseract, as better OCR provides a richer text signal for text-only models. Differences for image-based and multimodal models are small ($\leq$1.4~pp) and attributable to variation in hyperparameters, random seeds, or library versions.

Among zero-shot models on the original RVL-CDIP, vision-language models range from 42.26\% (CLIP) to 57.93\% (SigLIP~2), while zero-shot LLMs perform considerably better, ranging from 56.96\% (Mistral 7B Instruct) to 67.61\% (Llama 4 Maverick 17B). Both are well below the top supervised models but establish a useful baseline for interpreting cross-condition gains.

Turning to cross-condition comparisons, we observe several consistent patterns.
First, removing errors (Errors Dropped) results in a higher accuracy score for all models, highlighting the impact of data quality on performance evaluation.
Supervised models gain approximately 3 percentage points on average, while zero-shot LLMs see the biggest increases, each gaining roughly 5--6 points (e.g., gpt-oss-20b increases from 65.44\% to 71.08\%).

Second, when label errors are fixed (Errors Fixed), all models see a decrease in accuracy from the Errors Dropped scores.
This indicates that fixing the \emph{wrong} label error type exposes some overfitting by the models: since the label errors are pervasive across all folds of the dataset, a pervasively \emph{wrong} label error in the train set will lead the model to make ``correct'' predictions on erroneously labeled data. When those cases are fixed in the test set, the model will still predict the old label, which has now been corrected, resulting in misclassification.
When we address this issue by \emph{retraining} the model with corrected data (Fixed + Retrain), the accuracy scores increase again to levels above the Errors Fixed scores.
We see these Fixed + Retrain scores as representing what models achieve when label errors are addressed throughout RVL-CDIP.
The top-performing model here is LayoutLMv3, with a cleaned test accuracy of 97.32\%.

Finally, we observe that removing test samples with a (near-) duplicate in the train set leads to decreases in classification accuracy scores.
These drops range from approximately 1.8 to 3.6 percentage points across supervised models, with CNN models tending toward the higher end.
The zero-shot models are less impacted by data deduplication in the test set, with results mixed --- some models see small drops (SigLIP~2: $-1.69$~pp, Llama 4 Maverick 17B: $-2.33$~pp) while others are unaffected or improve slightly (CLIP: $+0.43$~pp, ALIGN: $+0.12$~pp); we speculate this may be because the removed duplicate test samples tend to be ``easier''.

Across all conditions, we also note little manifestation of the tendency of model leaderboards to ``destabilize''~\cite{pervasive-label-errors-2021} upon the removal or remediation of label errors.
In our experiments, top-performing models on the original RVL-CDIP still tend to be top performers on cleaned versions of the dataset.

% SigLIP2 model is siglip2-so400m-patch16-512
\begin{table}
\centering
\caption{Model performance on RVL-CDIP-N. The Original column charts performance on models trained on the original RVL-CDIP training set, while the Fixed column represents models trained on the cleaned (errors fixed) data. Zero-shot models (CLIP, ALIGN, SigLIP, SigLIP~2, and LLMs) require no training and are therefore evaluated under the Original condition only.}
\label{tab:rvlcdipn_performance}
\scalebox{0.85}{
\begin{tabular}{lccc}
\toprule
\textbf{Model} & \textbf{Original} & \textbf{Fixed} & $\Delta$ \\
\midrule
BERT & 88.12 & 91.32 & +3.20 \\
RoBERTa & 89.82 & 94.91 & +5.09 \\
LongFormer & 90.02 & 93.91 & +3.89 \\
Big Bird & 89.22 & 93.51 & +4.29 \\
\hline
VGG-16 & 66.97 & 80.34 & +13.37 \\
ResNet-50 & 62.08 & 75.55 & +13.47 \\
ResNeXT-50 & 65.17 & 77.94 & +12.77 \\
GoogLeNet & 60.38 & 67.47 & +7.09 \\
AlexNet & 62.08 & 72.85 & +10.77 \\
DocXclassifier & 74.95 & --- & --- \\
DiT & 78.84 & --- & --- \\
Donut & 78.84 & --- & --- \\
\hline
LayoutLM & 86.73 & 88.52 & +1.79 \\
LayoutLMv3 & 81.34 & 95.11 & +13.77 \\
\hline
CLIP & 81.64 & --- & --- \\
ALIGN & 89.22 & --- & --- \\
SigLIP & 92.51 & --- & --- \\
SigLIP 2 & 92.61 & --- & --- \\
\hline
gpt-oss-20b & 84.32 & ---& --- \\
%gpt-oss-120b & & & \\
Mistral 7B Instruct & 78.22 & --- & --- \\
Llama 4 Maverick 17B & 87.81 & --- & --- \\
Qwen 3 32b & 85.41 & --- & --- \\
\bottomrule
\end{tabular}}
\end{table}

\subsection{Out-of-Distribution Results}

Table~\ref{tab:rvlcdipn_performance} reports model performance on RVL-CDIP-N's out-of-distribution data. Every model trained on Fixed data outperforms its Original-trained counterpart, with the top Fixed-trained model (LayoutLMv3, 95.11\%) improving by +13.77~pp. We find that the magnitude of improvement splits along input modality: models that ingest pixels (the CNNs and LayoutLMv3) gain 7 to 14~pp, whereas OCR-based models (BERT, RoBERTa, and LayoutLM) gain less than 5.1~pp each. We interpret this as evidence that pixel-ingesting models were learning RVL-CDIP-specific visual shortcuts during Original training; training on corrected labels disrupts these shortcuts and encourages class-intrinsic features that transfer better out-of-distribution. OCR-token inputs are largely corpus-invariant, so text-based models have less to re-learn. 

Fixed training also substantially narrows the in-distribution-versus-OOD accuracy gap. Original-trained LayoutLMv3 drops from 94.30\% on RVL-CDIP to 81.34\% on RVL-CDIP-N, a gap of 13~pp; Fixed-trained LayoutLMv3 drops only from 97.32\% to 95.11\%, a gap of 2.2~pp. A similar narrowing holds for the CNN models. Label cleanup therefore does more than raise average accuracy---it restores a substantial portion of the OOD generalization that Original training otherwise sacrifices.

Notably, zero-shot models perform competitively on RVL-CDIP-N despite receiving no training on RVL-CDIP. SigLIP and SigLIP~2 achieve 92.51\% and 92.61\% respectively---outperforming every Original-trained supervised model, including LayoutLMv3 (81.34\%). Zero-shot LLMs (78--88\%) similarly match or exceed most pixel-ingesting supervised models trained on original data. This pattern is consistent with the RVL-CDIP shortcut interpretation: zero-shot models have never seen RVL-CDIP's training distribution and are therefore naturally unaffected by its domain-specific biases.

\section{Conclusion}

In this paper we conduct a thorough review of RVL-CDIP.
After a manual review, we find that approximately 12\% of the dataset is erroneously labeled.
We implement a filter-and-refine approach to detect test-train overlap in RVL-CDIP. In particular, we find that approximately 35\% of RVL-CDIP's test set has a near-duplicate or duplicate counterpart in the training set.
When we address these issues in new versions of RVL-CDIP, we find that model performance increases when label errors are addressed, but performance decreases if duplicates are removed from RVL-CDIP's test set.
We additionally find that training on error-corrected data substantially improves out-of-distribution generalization on RVL-CDIP-N, with models seeing improvements as large as 14 percentage points.
Our work brings to light several important data quality issues with the widely-used RVL-CDIP benchmark, calling attention to the need for high-quality and clean data.
We make our data and analyses available at: \url{https://rvlcdip-errors.com/}.

%\section*{Limitations}
% Limitations to our work include the possibility of human error in the manual review process.
% Additionally, we did not re-benchmark every model published using RVL-CDIP, as many do not have public implementations or are not straightforward to re-implement.
% Our duplicate detection method is evaluated specifically on RVL-CDIP; while we make no claims about its generalizability, the filter-and-refine pipeline is dataset-agnostic in principle and should be applicable to other document image datasets.
% Finally, our out-of-distribution evaluation is limited to RVL-CDIP-N~\cite{larson-etal-2022-rvlcdipn}, as it is the only published benchmark we are aware of that introduces genuinely new out-of-distribution document data for RVL-CDIP, rather than augmentations of existing RVL-CDIP samples.

%\begin{acks}
%The authors have no competing interests to declare that are relevant to the content of this article.
%\end{acks}

\bibliographystyle{ACM-Reference-Format}
\bibliography{custom_no_urls}
%\bibliography{custom}

\end{document}